\documentclass[times,twocolumn,final,10pt]{elsarticle}
\usepackage{prletters}
\usepackage{framed,multirow}
\usepackage{amssymb}
\usepackage{latexsym}
\usepackage{epsfig}
\usepackage{graphicx}
\usepackage{amsmath}
\usepackage{algorithm}
\usepackage{float}
\usepackage{array}
\usepackage{longtable}
\usepackage{setspace}
\usepackage{caption}
\usepackage{enumerate}
\usepackage{url}
\usepackage{soul}

\AtBeginDocument{%
	\abovedisplayskip=5pt plus 2pt minus 3pt
	\abovedisplayshortskip=2pt plus 3pt
	\belowdisplayskip=5pt plus 2pt minus 5pt
	\belowdisplayshortskip=2pt plus 3pt minus 3pt
}

\usepackage{url}

\journal{Pattern Recognition Letters}

\begin{document}
	
	\begin{frontmatter}			
		
		\title{An Efficient Ensemble Explainable AI (XAI) Approach for Morphed Face Detection}
  
		\author[1]{Rudresh \snm{Dwivedi}\corref{cor1}} 
		\cortext[cor1]{Corresponding author: 
			Tel.: +91-971-388-8726;}
		\ead{rudresh.dwivedi@nsut.ac.in}
				 \author[1]{Ritesh  \snm{Kumar}}\ead{ritesh.cs19@nsut.ac.in}
                 \author[1]{Deepak  \snm{Chopra}}\ead{deepak.chopra.cs19@nsut.ac.in}
		   		 \author[1]{Pranay \snm{Kothari}}\ead{pranay.cs19@nsut.ac.in}
                 \author[1]{Manjot  \snm{Singh}}\ead{manjot.cs19@nsut.ac.in}
             		
		\address[1]{Department of Computer Science and Engineering, Netaji Subhas University of Technology, Delhi-110078, India}
		
		\communicated{Rudresh Dwivedi}

\begin{abstract}
	The extensive utilization of biometric authentication systems have emanated attackers / imposters to forge user identity based on morphed images. In this attack, a synthetic image is produced and merged with genuine. Next, the resultant image is user for authentication. Numerous deep neural convolutional architectures have been proposed in literature for face Morphing Attack Detection (MADs) to prevent such attacks and lessen the risks associated with them. Although, deep learning models achieved optimal results in terms of performance, it is difficult to understand and analyse these networks since they are black box/opaque in nature. As a consequence, incorrect judgments may be made. There is, however, a dearth of literature that explains decision-making methods of black box deep learning models for biometric Presentation Attack Detection (PADs) or MADs that can aid the biometric community to have trust in deep learning-based biometric systems for identification and authentication in various security applications such as border control, criminal database establishment etc. In this work, we present a novel visual explanation approach named Ensemble XAI integrating Saliency maps, Class Activation Maps (CAM) and Gradient-CAM (Grad-CAM) to provide a more comprehensive visual explanation for a deep learning prognostic model (EfficientNet-B1) that we have employed to predict whether the input presented to a biometric authentication system is morphed or genuine. The experimentations have been performed on three publicly available datasets namely Face Research Lab London Set, Wide Multi-Channel Presentation Attack (WMCA), and Makeup Induced Face Spoofing (MIFS). The experimental evaluations affirms that the resultant visual explanations highlight more fine-grained details of image features/areas focused by EfficientNet-B1 to reach decisions along with appropriate reasoning. 
\end{abstract}

\end{frontmatter}

\section{Introduction}
\subsection{Background}
The indispensable growth of Artificial Intelligent (AI) systems as evidenced by superior Machine Learning (ML) systems outperforming conventional handcrafted techniques across many application areas is primarily due to advancements in deep learning techniques, availability of powerful Graphical Processing Units (GPUs) with high computational gains and access to large datasets as well \cite{lecun2015deep,karpathy2014large}. Currently, a major concern with AI systems is the lack of transparency of Deep Learning algorithms \cite{samek2017explainable,doshi2017towards,holzinger2019causability,emmert2020explainable}. Nonetheless, neural networks are known to have optimal performance, the research community is conducting in-depth studies to comprehend the limitations of black box algorithms and the significance of understanding the insights behind a neural network's decisions and choices made by them. Incorporating interpretability in biometric authentication systems may have a conducive effect. While designing a PAD scheme, instead of just anticipating a model's outputs, it will be more insightful to find the rationale behind its output predictions. It is of paramount importance to integrate interpretability techniques to evaluate and understand the predictions of biometrics PAD predictions made by black box algorithms. Furthermore, conventional metrics quantify a model's performance by relying primarily on the predicted labels without investigating the data that has been utilised to make these conclusions. Such evaluation is extremely constrained for deep learning-based approaches in particular.

Based upon the above mentioned limitations in present state-of-the-art \cite{rec1,rec2,rec3,samek2017explainable,doshi2017towards,holzinger2019causability,emmert2020explainable}, the prime focus of our work is on reformulating the evaluation framework so as to be more detailed, understandable and meaningful. This goal can be achieved through interpretability as it leverages complementary insights about the complex operations of deep learning models.
Additionally, through interpretability, it is possible to analyze their behaviour for unseen attacks and how robust a model is at generalizing to unseen attacks.

\subsection{Motivation and Contribution}
This work explores the need and prospects of interpretability  and explainability in determining a model's effectiveness in biometric PAD for face modality. The key contributions of this work are as follows:
\begin{itemize}

    \item In this work, we have deployed an efficient deep neural architecture i.e EfficientNet-B1 as a PAD approach to distinguish a morphed image from a genuine one. 
    \item Additionally, we have used TensorBoard, a visualization tool provided by TensorFlow to visualize the output of an intermediate layer of the EfficientNet-B1 model. This simply lets us visually determine how the output was conceived by looking at each layer, although it doesn't give us concise information, resulting in the need for gradient-based methods that we have used further.
    \item Next, we propose a novel ensemble to produce visual explanations by combining three interpretability methods namely saliency maps, CAM and Grad-CAM to develop an optimal interpretable framework.
    \item The EfficientNet-B1 model is trained and evaluated separately on three datasets namely Face Research Lab London Set, Wide Multi-Channel Presentation Attack(WMCA) Set and Makeup Induced Face Spoofing(MIFS) Set. The ensemble XAI approach has been implemented on all three datasets mentioned above. Finally, we have performed a rigorous experimentation for the influence of visual explainability in biometric PAD considering the pros and cons of the three interpretability methods considered in our work.
\end{itemize}

\subsection{Interpretability Approaches}
\begin{figure}[h]
\includegraphics[width= 10cm]{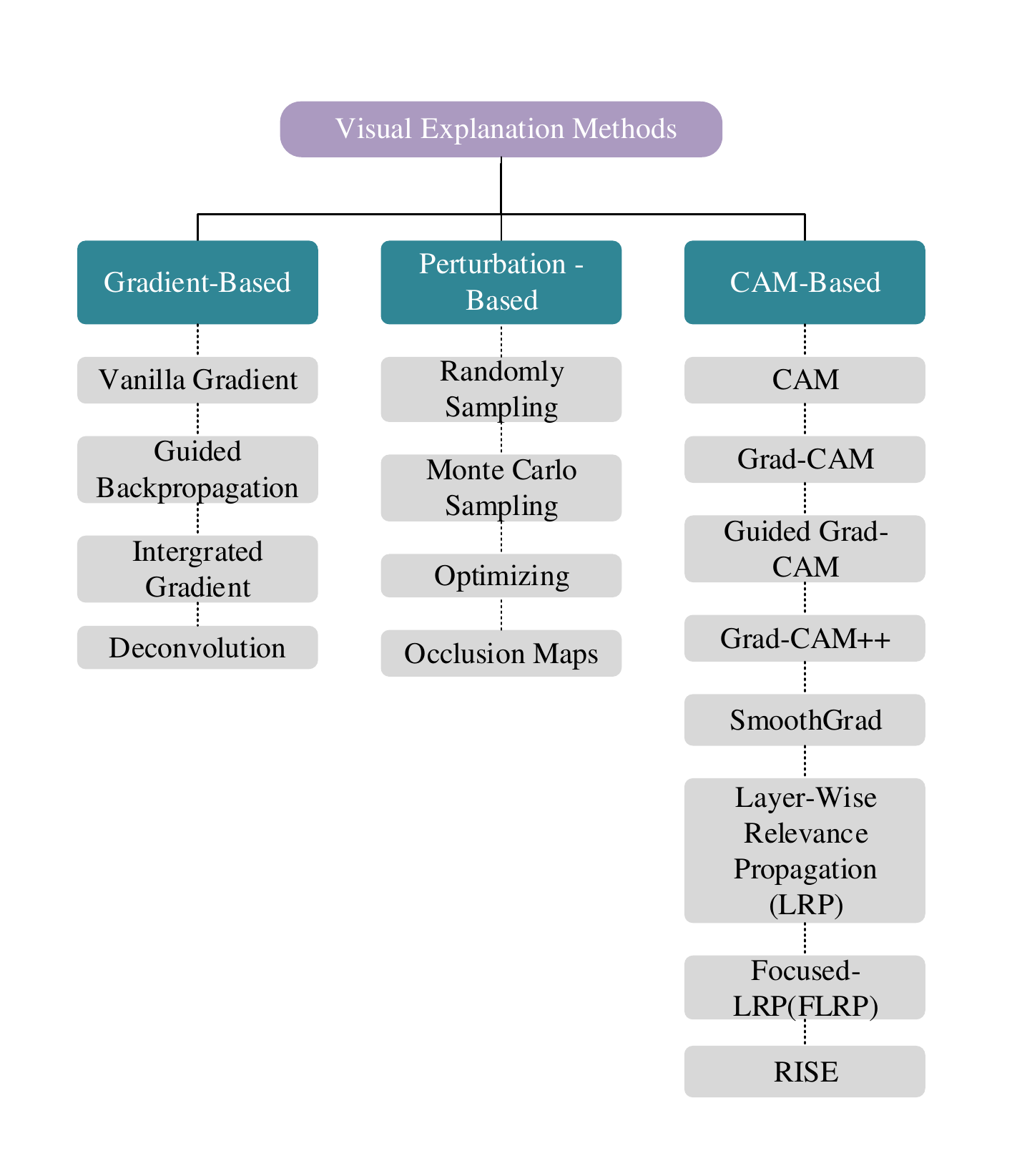}
\caption{A taxonomy of Visual Explanation Methods}
\label{F1}
\end{figure}

In literature, there exists three different schemes of visual explanations as illustrated in Fig \ref{F1}. First of them, Gradient-based \cite{simonyan2014very} visualizations utilize the back-propagated gradient to generate a sensitivity map in input space. Next, perturbation-based approaches \cite{ribeiro2016should} measure the feature importance by masking. Lastly, the CAM-based \cite{zhou2016learning} schemes generate  interpretable attribution maps that highlight the prominent regions in the target image and have been adopted in many tasks. These approaches primarily answer to a very important question such as \textit{ `Can we know what a network was looking at while predicting a class?'.}

In this section, we have discussed three state-of-the-art interpretability or visual explanation approaches targeted for our work followed by introducing the Ensemble XAI approach.

\subsubsection{Saliency Maps}

Saliency maps \cite{simonyan2013deep} are the most frequently used explanation method for interpreting the decisions of CNNs. It indicates the areas of the image that gets affected upon the deployment of each layer of network. The authors in \cite{zeiler2014visualizing} introduced deconvolutional networks to understand what an intermediate layer of a network is looking at (features or pixels) so as to reconstruct the original image from the activations of that intermediate layer. Using a back-propagation algorithm to determine the gradients of logits for the input image fed to the network, the authors in \cite{simonyan2013deep} developed a further simple method for obtaining saliency maps. Through back-propagation, one can identify those pixels in the input image that contribute the most to the final score by highlighting them according to the gradient they receive. Springenberg et al. \cite{springenberg2014striving} proposed the guided back-propagation algorithm as the third approach for generating saliency maps which are quite helpful in generating sharp and high-resolution saliency maps. However, few research studies say that saliency maps are not always reliable \cite{kindermans2019reliability}, \cite{ghorbani2019interpretation}.

\subsubsection{Class Activation Maps (CAM)}

CAMs are another state-of-the-art that provides explanations in CNNs \cite{zhou2016learning,zhu2017soft}. A deep CNN's final stack of completely linked layers is not used in CAM; instead, the concept of Global Average Pooling (GAP) is used. Considering only the last convolutional layer and then for each feature map in that convolutional layer, GAP is employed. By reflecting the output weights on the feature maps of the final convolutional layer, this architecture highlights the salient areas of an image. The weighted sum of all these activation maps provides the contribution of each towards one particular class label. This approach yields a way of getting class discriminative saliency maps. The key feature of CAM lies in it's class discriminability. Hence, it can localize objects without positional supervision. There are few limitations associated with CAM as well, the key one is the need of retraining the trained models for explanation leading to additional computational overhead. Also, there is a constraint on architecture that it is necessary to introduce a GAP layer to be able to explain the model and that may degrade the performance. The class score for CAM is described as follows:

\begin{equation}
Y^{c} = \underset{k}{\sum}\underset{\substack{\text{Class feature} \\ \text{weights}}}{\underbrace{w_{c}^{k}}}\overset{\substack{\text{Global avg.} \\ \text{pooling}}}{\overbrace{\frac{1}{z}\underset{i}{\sum}\underset{j}{\sum}}}\underset{\substack{\text{feature} \\ \text{map}}}{{\underbrace{A_{k}^{ij}}}}
\end{equation}
where,  \({Y^{c}}\) is the class scores in the last layer, \(A_{k}^{ij}\) is the pixel at (i,j) location of \(K^{th}\) feature map.

\subsubsection{Grad-CAM}

Grad-CAM viz. Gradient-Weighted CAM \cite{selvaraju2017grad} is an approach to re-purposing CAM using existing quantities in a CNN and gained much popularity in recent years. In Grad-CAM, the ReLu activation function is applied on the weighted combination of activation maps to ensure only positive correlations are being shown in final saliency maps. The feature map's importance is evaluated as follows:

\begin{equation}
\alpha _{k}^{c} = \overset{\substack{\text{Global avg.} \\ \text{pooling}}}{\overbrace{\frac{1}{z}\underset{i}{\sum} \underset{j}{\sum}}}\underset{\substack{\text{gradients} \\ \text{via backprop}}}{\underbrace{\frac{\partial Y_{c}}{\partial A_{k}^{ij}}}} 
\end{equation}
where $c$ is the class of interest and $k$ is the index of the activation map of the final convolutional layer. The above-evaluated $\alpha _{k}^{c}$ indicates the significance of feature map $k$ for the intended class $c$. The values are then added together after multiplying each activation map by its importance score (i.e. $\alpha _{k}^{c}$). Finally, the ReLu non-linearity is applied to the summation in order to find GradCAM heatmap to class of interest. The GradCAM heatmap ($L_{Grad-CAM}^{c}$) is defined as:

\begin{equation}
L_{Grad-CAM}^{c} = ReLu \underset{\substack{\text{linear} \\ \text{combination}}}{\underbrace{\underset{k}{(\sum }\alpha _{k}^{c}A^{k})}} 
\end{equation}

\subsection{Ensemble Learning}
\begin{figure*}[!htbp]
	\centering
	\includegraphics[width=\textwidth]{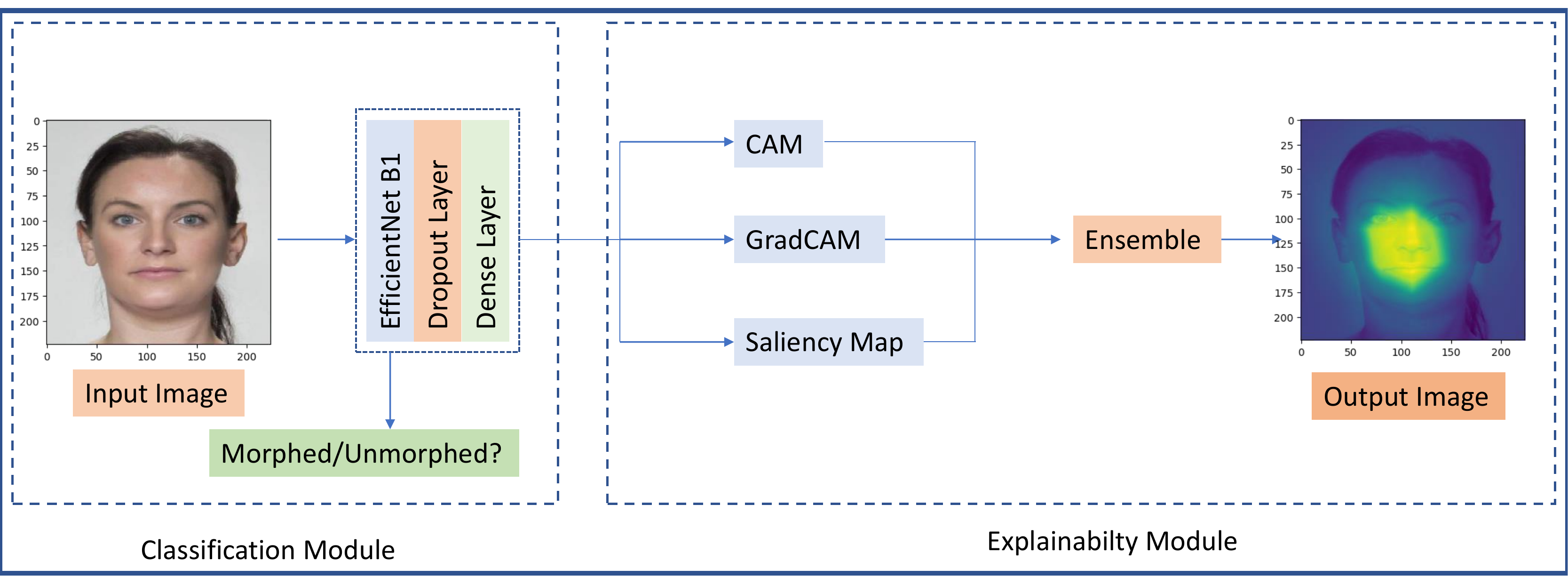}
	\caption{Work Flow of the proposed approach}
	\label{proposed}
\end{figure*}

Ensemble learning integrates the salient aspects of different models to obtain a shared consensus for the model's predictions. The framework are designed to reduce variation in prediction errors, making it more reliable than the individual models that make up the ensemble. Also, it tends to fetch complementary information through the impact from various models. Ensemble techniques are divided into three categories namely stacking, bagging and boosting. In our work, we have used the stacking ensemble technique for combining the complementary information provided by saliency maps, CAM and Grad-CAM. The block diagram of the proposed ensemble XAI framework is illustrated in Fig. \ref{proposed}.

The remainder of this work is organised as follows. Section II summarises the related work on interpretable face biometrics. Section III outlines the proposed work. In section IV we have discussed experimental results and analysis. Section V summarises the study's findings and provides recommendations for future research into more sophisticated interpretability tools for biometrics PAD. 

\section{Related Work}
\subsection{Interpretability and Explainability: Concept and Literature}
Over the past few years, the research community have put in a lot of effort in developing both interpretable models (pre‐model and in‐model stage) and explanation methods (post‐model stage) as well \cite{doshi2017towards,samek2017explainable,kim2018interpretability,silva2018towards,silva2019produce}.
Nevertheless, few works highlight the distinctions between these two concepts. Doshi-Velez et al., in their study \cite{doshi2017towards}, define interpretability as “the ability to explain or to present in understandable terms to a
human”. Miller \cite{miller2019explanation} defines the term interpretability as “the degree to which a human can understand the cause of a decision”. Based on the above definitions, it can be concluded that interpretability is primarily related to the rationale behind a model's outputs. 

Kim et al. \cite{kim2018interpretability} argue that interpretability is an essential component of building trustworthy and reliable machine learning systems. However, explainability signifies the internal reasoning and mechanics of a machine learning model. As the model becomes more explainable, humans tend to achieve a deep understanding of the model in terms of internal procedures being carried out while the model is learning or drawing conclusions. The authors in \cite{silva2018towards,silva2019produce} highlight the importance of explainability where decisions based on predictions can have significant consequences and argue that simply providing a single explanation for a model's prediction may not be sufficient to build trust in the model or to identify potential biases. A model that is interpretable does not always imply that individuals can comprehend its underlying logic or operations. As a result, interpretability for any model does not invariably or intuitively imply explainability and vice versa. Gilpin et al. \cite{gilpin2018explaining} provided support for the idea that explainability is equally important to interpretability and that the absence of either is unacceptable and also concluded in his study that interpretability is a much broader term than explainability. A substantial amount of research is being done in the literature to interpret and explain how machine learning systems behave in different applications highlighting the benefits and need for transparent decision-making.

\subsection{XAI in the field of biometrics and PAD}

Many predominant works contributed by various authors in the literature highlight the fact that the use of efficient deep CNN-based architectures for PAD for face modalities, in particular, has demonstrated remarkable performance in terms of performance accuracy.
Recent works have also addressed the issue of PAD generalization \cite{gilpin2018explaining,george2020learning,perera2019learning,xiong2018unknown}, thereby enhancing the resistance of face PAD systems to unseen attacks. The authors in \cite{gilpin2018explaining} provide an overview of the different approaches to interpretability in machine learning, such as rule-based methods, visualization techniques and model-agnostic methods. Patel et al. \cite{perera2019learning} proposed a novel deep learning framework for one-class classification, called Deep SVDD (Support Vector Data Description), which learns a high-dimensional data representation using a deep neural network. Zee et al. \cite{zee2019enhancing}  presented in their work the interpretability potential of a Siamese CNN to assist humans in hard prediction tasks i.e. authors used (CAM) to primarily know where to look in the subject images to correctly understand the decisions made by black box CNN models. In \cite{liu2018learning}, the authors used the depth map and the Remote Photoplethysmography (rPPG) signal as the auxiliary supervision to enhance the effectiveness of the face anti-spoofing method using a Residual Neural Network. In recent years, authors are trying to exploit the methodologies of explainability and interpretability of a ML models in the field of biometrics, in particular, face PAD by making attempts such as depth map \cite {wang2020deep} \cite{yu2020searching}, \cite{shao2020regularized}, producing saliency maps for CNN models \cite{liu2020disentangling}, \cite{yang2019face} or some relevant works on estimating the patterns that define a spoofed sample \cite{liu2020disentangling}, \cite{jourabloo2018face}. The authors in \cite {wang2020deep} proposed a method that includes two sub-networks, a spatial network, and a temporal network. The spatial network processes the input image to extract spatial features, while the temporal network processes a sequence of images to capture temporal features. In \cite{yu2020searching}, the authors proposed a Central Difference Convolutional Network (CDCN) architecture that is specifically designed for this task. The CDCN is a deep neural network that accepts a face image as input and outputs a probability score indicating whether the input image is genuine or spoofed. The authors in \cite{shao2020regularized} proposed an approach consisting of two stages: the meta-training stage and the meta-testing stage. In the meta-training stage, a meta-learner is trained to learn a good initialization for the feature extractor network that is used to extract features from face images. In the meta-testing stage, fine-tuning is performed on a smaller dataset of the same type as the one used in the meta-training stage. In \cite{liu2020disentangling}, the authors introduced a method that disentangles the spoof trace from the real face signal by exploiting the differences in the optical flow patterns between real and fake faces, leading to an effective and robust generic face anti-spoofing system. The authors in \cite{phillips2020four} outlined four XAI principles for biometrics and face recognition fields to build trust and fulfil societal norms. Seilbold et al. \cite{seibold2021focused} worked on the limitation faced by LRP \cite{bach2015pixel} by proposing a new approach named Focused Layer-wise Relevance Propagation (FLRP) which is an extension of LRP to assist humans in inspecting morphed face images. The authors emphasized precise focus on pixel levels to determine what images regions a particular DNN uses for distinguishing between morphed or genuine image. To accomplish this the authors presented that the key idea behind FLRP is to concentrate on the learnt features that characterize a morphed face image by starting the relevance propagation from neurons in an intermediate layer rather than those in the final one as is the case in normal LRP. Moreover, the authors proposed a novel framework for evaluating interpretability methods for various morphing attack detectors. To the best of our knowledge so far, our work is the first attempt to work in the direction of ensemble-based interpretability approaches for enhanced visual explanations for deep CNNs.

\section{Proposed Approach}
The proposed work aims to develop an ensemble-based interpretability approach to support the decisions of Efficient-B1 for distinguishing between morphed and unmorphed face images by providing enhanced visual explanations. In this section, we present the procedural steps to acquire model explanations.

\subsection{Pre-processing}
Real-world data usually contains ambiguous formats, missing values, and noise that a ML models cannot directly process. Hence, pre-processing is required to mitigate these infelicities. In our work, first, we resize the images in $224 \times 224$ to feed the data into network. For WMCA dataset, the images were available in HDF5 format. It was required to be converted into JPEG frames for training. Next, the  dataset is then further split into training and test data (approximately in an 80:20 ratio) for all three datasets. 

\subsection{Model Selection and Training: EfficientNet}
In 2019, Google published a study that showcased a novel family of CNNs i.e EfficientNet \cite{tan2019efficientnet} aiming to provide efficacy in terms of performance, number of parameters, and FLOPS (Floating Point Operations Per Second). First, an elementary mobile-sized baseline architecture i.e. EfficientNet-B0 is applied that achieved 77.3\% accuracy on ImageNet with $5.3$M parameters and $0.39$ billon FLOPS. The total number of various layers in a network is determined by the \textit{depth (d)} of the network. The \textit{width (w)} determines how many filters or neurons there are in a convolutional layer. Thereafter, the authors proposed a novel scaling technique i.e. compound scaling for increasing the model size by scaling all the three dimensions (depth, width and resolution) uniformly to achieve maximum accuracy gains. The key idea behind the compound scaling method was to use a compound coefficient $\phi$ to uniformly scale all aforementioned network's dimensions in a more principled way thus deriving a series of eight models i.e. EfficientNetB0-B7. These parameters can be evaluated as follows:

\begin{equation}
\begin{aligned}
depth: d &= \alpha^{\phi} \\
width: w &= \beta^{\phi}  \\
resolution: r &= \gamma^{\phi}  \label{dwr}
\end{aligned}
\end{equation}

such that:

\begin{equation}
\begin{aligned}
\alpha. \beta^{2}.\gamma^{2} \approx 2    \\
 \alpha \geq 1, \beta \geq 1, \gamma \geq 1
\end{aligned}
\end{equation}
here, \({\alpha, \beta, \gamma}\) are determined by applying a small grid search.\(\Phi\) is a user-specified coefficient that determines the number of resources available for model scaling. The constants \(\alpha, \beta, \gamma\) specify the ways through which these extra resources can be allocated to a network's width, depth, and resolution respectively.

\begin{figure}[htbp]
	\includegraphics[width=85mm,scale=0.7]{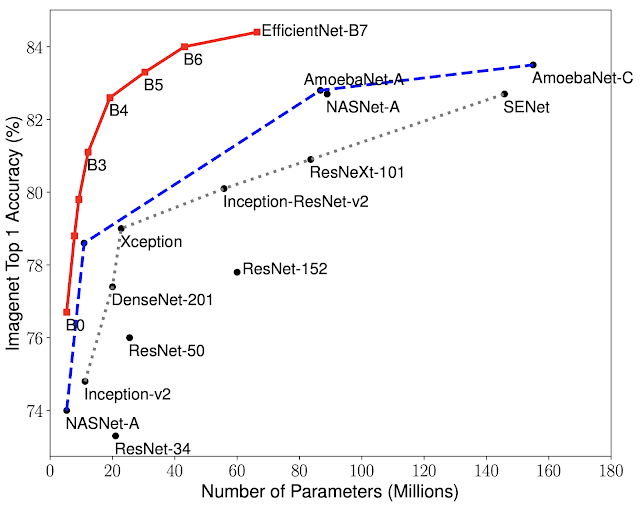}
	\caption{Model size vs. accuracy comparison \cite{tan2019efficientnet}. EfficientNet-B0 is the baseline network developed by AutoML MNAS, while Efficient-B1 to B7 are obtained by scaling up the baseline network.}
	\label{f31}
\end{figure}

\begin{figure}[htbp]
    \includegraphics[width=85mm,scale=0.7]{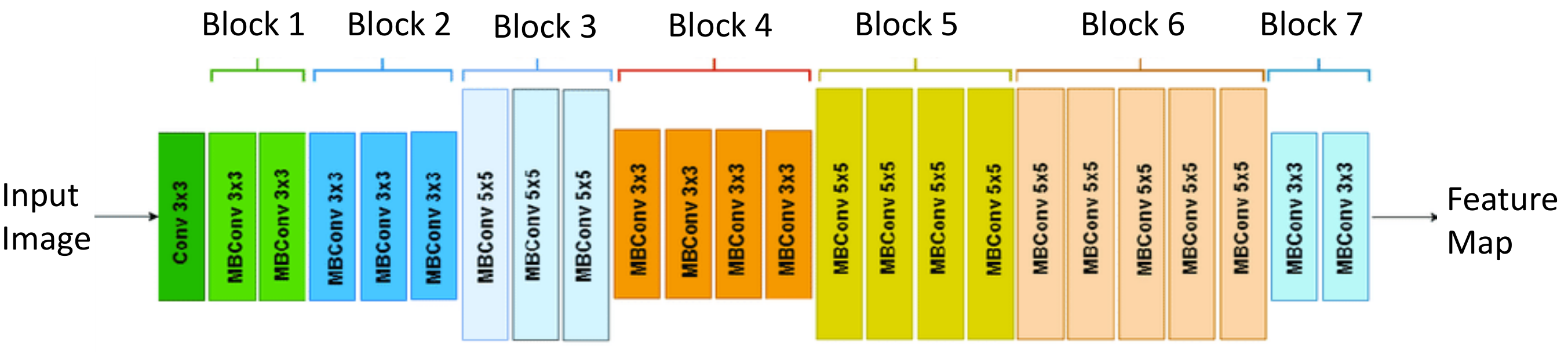}
    \caption{Model Architecture of EfficientNet-B1 with Inverted Residual Block (MBConv) as basic building block.}
    \label{f3}
\end{figure}

For our work, we have employed EfficientNet-B1 as a classifier as it outperforms other CNNs (see Fig. \ref{f31}). Also, it involves fewer parameters in comparison to other models such as VGG16, VGG19, ResNet etc. The model architecture of EfficientNet-B1 is shown in Fig. \ref{f3}.

	\subsection{Domain Specific Learning}

	In this work, we have used a well-known pre-trained model i.e EfficientNet-B1 as the base network and then fine-tuned it by adding few custom layers as per our problem statement. On top of the EfficientNetB1, we used a dropout layer preventing overfitting and an output layer to enhance the model efficacy. The network accepts as input an image of size 224 x 224. The custom block consists of GAP layers with a dropout rate of 0.5. The proposed model shows promising results for detecting morphed and unmorphed images as discussed in result metrics in Section \ref{expt}.
	
	\subsection{Model Explanation using Ensemble}
	The proposed model is deployed over images to generate heatmaps. These heatmaps are then fed to three different functions for CAM, GradCAM, and Saliency Map respectively. The produced results of these functions is then superimposed to produce an ensemble by stacking. The ensemble model thus produced is then used to assess which features of the face have an impact on morphed/unmorphed classification. Additionally, TensorBoard, a powerful visualization tool provided by TensorFlow is used to visualize the output of a layer using callbacks after an epoch to comprehend the working insight of a particular layer of EfficientNet-B1. This allowed us to determine which features were being used to make further improvements in the weights of the neural network.

\section{Experimental Results and Analysis} \label{expt}

To provide model explanations using ensemble, we perform a number of experiments to demonstrate the efficacy of our method. Section \ref{dataset} describes the databases used in our work. Section \ref{setup} narrates the experimental setup employed in our work. Next, the performance metrics to evaluate performance are defined in Section \ref{metric}. Furthermore,
the we discuss the performance results in Section \ref{results}. Finally, TensorBoard based visualizations are provided in Section \ref{tensor}.

\subsection{Datasets}\label{dataset}
In Our work, we have utilized three different datasets for experimental evaluations. The details of these datasets are as follows:

\noindent \textbf{Face Research Lab London (FRLL) dataset}: FRLL dataset \cite{londondata} is a well-known datasets available for all categories of face-processing and analysis tasks. This dataset comprises of 102 faces which incudes 49 females and 53 males. (age Mean = 27.72, Standard Deviation = 7.11, Range = 18-54). 

\noindent \textbf{Wide Multi Channel Presentation Attack (WMCA) dataset}: WMCA dataset \cite{george2019biometric} is formed by capturing 1941 short video clips of both legitimate and presentation attacks from 72 individuals. The data is being recorded through multiple channels such as colour, depth, infrared and thermal during four sessions with varying  environmental conditions. Different categories of presentation attacks included in this dataset are bonafide, glasses, fake head, print, replay, wig, makeup and also masks such as rigid, flexible or even paper masks. 

\noindent \textbf{Makeup Induced Face Spoofing (MIFS) dataset}: The prime motive behind acquisition of MIFS dataset \cite{chen2017spoofing} is to investigate spoofing factors caused by makeup. A total of 107 makeup transformations are captured randomly from makeup tutorials available on YouTube. The MIFS dataset includes images of the target subject being spoofed as well as images of the subject being spoofed before and after makeup (the same subject). The target subject images aren't always the ones that the YouTube user used as a reference while applying makeup transformations.

\subsection{Implementation Setup} \label{setup}

In our work, we have utilized a renowned transfer learning model i.e EfficientNet-B1 trained on the ImageNet dataset as the base model and fine-tuned it by adding a dropout and output layer for our proposed work. For training purpose, the batch size and number of epochs are considered as 32 and 5, respectively. We have trained and evaluated the performance of EfficientNet-B1 on all three datasets separately. All the datasets are split in 80:20 ratio for training and evaluation, and classification among morphed and unmorphed images.

Next, we have deployed a stacking-based ensemble approach which utilises the output of base interpretation method. Due to their capacity to reduce bias and variation, ensemble methods are frequently used in deep networks. The gradient-based XAI methods i.e. Saliency maps, CAM, and Grad-CAM are applied simultaneously to generate visual explanations. The model produces three heat maps: CAM, Grad-CAM, and Saliency Map for each input image.  Next, the regression is applied to the normalised Grad-CAM, CAM, and Saliency Map outputs to identify discriminative regions and target labels. Different intensity regions are present on the target label. All three Grad-CAM, CAM, and Saliency Map heatmaps are then scaled and transformed into a 1-D features array. Grad-CAM, CAM, and Saliency map pixel arrays are then concatenated to create a single, combined pixel array, which creates a ensemble of the gradients of all three approaches. 
\subsection{Evaluation Metrics} \label{metric}

We have used the following performance matrices to measure the performance of EfficientNet-B1 for PAD on all three datasets:

\begin{itemize}
\item \textbf{Accuracy}: The ratio of the overall number of samples that were correctly predicted to the overall sample count is used to determine the classifier's accuracy.
\begin{equation}
{Accuracy = \frac{TP+TN}{TP+TN+FP+FN}}  
\end{equation}  
where,

True Positives (TP) are situations in which the predicted ``Yes" truly belonged to the ``Yes" class.
True Negatives (TN) are situations in which the predicted ``No" truly belonged to the category ``No".
False Positives (FP) are situations where the predicted ``Yes" was instead a member of the class ``No".
False Negatives (FN) are situations where the predicted ``No" actually belonged to the ``Yes" class.
\item \textbf{Precision}: True positives (TP) divided by the total of true positives (TP) and False positives(FP) is known as the Precision.
\begin{equation}
{Precision = \frac{TP}{TP+FP}}
\end{equation} 
\item \textbf{Recall}: The recall is the ratio of true positives (TP) by the sum of true positives (TP) and false negatives (FN).
\begin{equation}
{Recall = \frac{TP}{TP+FN}}
\end{equation} 
\item \textbf{F1 score}: When accuracy and recall are significant to the use case, the F1 score should be employed. The harmonized averaged accuracy and recall make up the F1 score. Its range is [0, 1].
\begin{equation}
{F1 score = 2\ast \frac{Precision*Recall}{Precision+Recall}}
\end{equation} 
\end{itemize}
Apart from the above metrics, we have used some metrics defined by the ISO standards \cite{isoap} to measure the performance of the proposed PAD algorithm:
\begin{itemize}
    \item \textbf{Attack Presentation Classification Error Rate (APCER)}: It is defined as the rate of mis-classified spoof images (Spoof predicted as live). 
    \begin{equation}
{APCER=1.0*\frac{FN}{FN+TP}}   
\end{equation}    
\end{itemize}
\begin{itemize}
    \item \textbf{Bonafide Presentation Classification Error Rate (BPCER)}: It is defined as the rate of misclassified live images (Live predicted as spoof). 
\begin{equation}
{BPCER=1.0*\frac{FP}{TN+FP}}       
\end{equation}    
\end{itemize}
\begin{itemize}
    \item \textbf{Half Total Error Rate (HTER)}: It is defined as the average of the False Acceptance Rate (APCER) and False Rejection Rate (BPCER). 
\begin{equation}
{HTER = 1.0*(APCER + BPCER)/2}
\end{equation}    
\end{itemize}     

\subsection{Results and discussion}\label{results}

In this section, we discuss the performance of our proposed model for PAD as well as the impact of ensemble XAI to further ensure the reasons behind a model's particular prediction by having the heatmaps generated  through our proposed Ensemble XAI model of interpretability methods for image dataset.
\begin{enumerate}
\item \textbf{Face Research Lab London Dataset}: For training data, the accuracy of the model observed  is 98.05\%. The loss observed is 0.0715. Precision is evaluated as 97.84\%. The recall is evaluated as 1.0. The F1 score is calculated as 0.9891. For test data, the accuracy of the model observed  is 96.76\%. The loss observed is 0.1673. Precision is evaluated as 96.39\%. The recall is evaluated as 1.0. The F1 score is calculated as 0.9816. 

\item \textbf{MIFS Dataset}: For training data, the accuracy of the model observed  is 66\%. The loss observed is 1.5277. Precision is evaluated as 100\%. The recall is evaluated as 0.3131. The F1 score is calculated as 0.4769. For test data, the accuracy of the model observed  is 65.42\%. The loss observed is 2.2155. Precision is evaluated as 100\%. The recall is evaluated as 0.3157. The F1 score is calculated as 0.4799.

\item \textbf{WMCA Dataset}: For training data, the accuracy of the model observed  is 100\%. The loss observed is 0. Precision is evaluated as 100\%. The recall is evaluated as 1. The F1 score is calculated as 1. 
For test data, the accuracy of the model observed  is 100\%. The loss observed is 0. Precision is evaluated as 100\%. The recall is evaluated as 1. The F1 score is calculated as 1.
Table \ref{baa} summarizes the performance of our proposed model on all three datasets.
\begin{table*}[!htbp]
\caption{Performance of the proposed model}
\label{baa}
\centering
\begin{tabular}{|l|l|l|l|l|l|l|l|}
	\hline
	Datasets used & Accuracy (\%) & Precision & Recall & F1 Score & APCER  & BPCER  & HTER   \\ \hline
	FRLL          & 96.76         & 0.96      & 1.0    & 0.9816   & 0      & 0.2419 & 0.1209 \\ \hline
	MIFS          & 65.42         & 1.0       & 0.3157 & 0.47     & 0.6842 & 0      & 0.3421 \\ \hline
	WMCA          & 100           & 1.0       & 1.0    & 1.0      & 0.0098 & 0      & 0.0049 \\ \hline
\end{tabular}
\end{table*}

\textbf{Visualization analysis:}
In this section, we analyse the impact of applying an Ensemble of three state-of-the-art interpretable approaches to understand the reasoning behind our proposed model's prediction through visual explanations in the form of heatmaps. As shown in Fig. \ref{Face}, Fig. \ref{MIFS} and Fig. \ref{WMCA}. firstly, we have generated the heatmaps by applying saliency, CAM and Grad-CAM on morphed and unmorphed images from all three datasets separately. Saliency maps are describing the salient pixel attributions in the input image that the model is referring to in predicting a class label as morphed or unmorphed. Similarly, when we applied Grad-CAM on the input image, the heatmap generated is showing class discriminative features to classify the target image and the application of Grad-CAM on the target image, which is an improvement over CAM generates heatmaps showing areas with different colour intensities. The red region describes the most prominent features, whereas, the blue region describes the least preferable regions considered by the model for predicting a particular class label as morphed or unmorphed.

We have superimposed saliency maps, CAM and Grad-CAM heatmaps in a stacking manner on the input image to make an ensemble XAI as shown in Fig. \ref{Face}, Fig. \ref{MIFS}, and Fig. \ref{WMCA} respectively. The visualization exhibit that the network trained purely for face morph detection focused more on the periocular region and on the peri-nose region but less on the mouth region, which resulted in the discrimination of this area as indecisive. The heatmaps also depicts that the neural network has learned distinct deep features for all regions of the image affirming the robustness of the architecture. The ensemble, in turn, has identified prevalent regions and provided insight into the decisions.

\begin{figure*}[!htbp]
	\centering
	\includegraphics[width=0.89\textwidth]{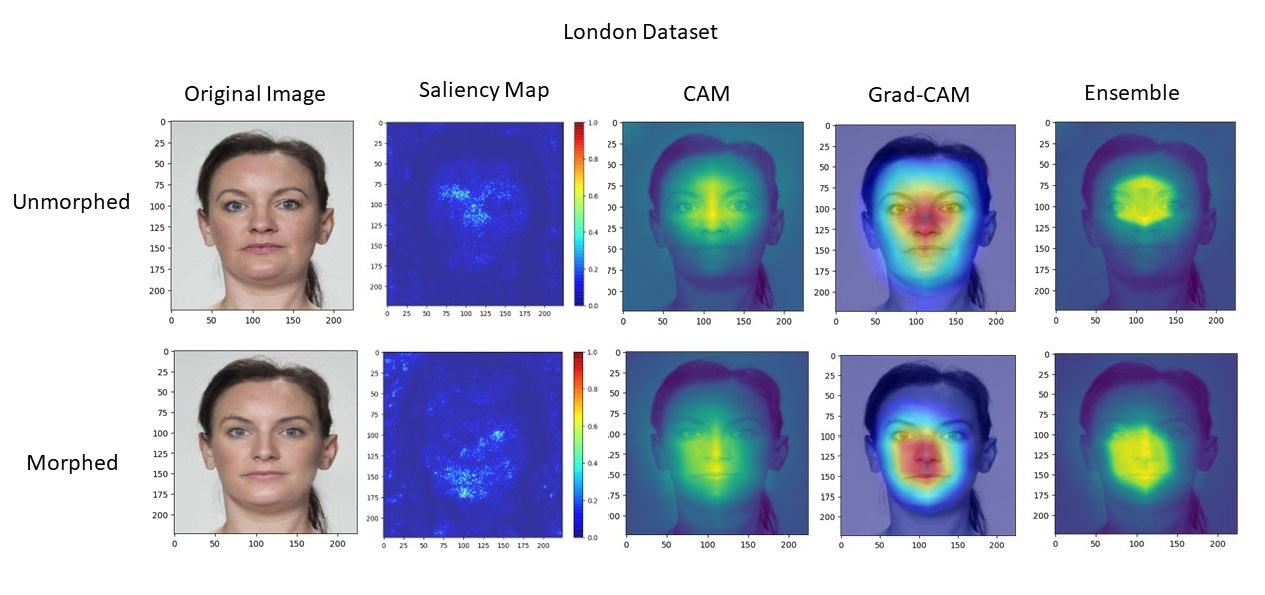}
	\vspace{-6mm}
	\caption{The salient regions for categorising the image as morphed/unmorphed for Face Research Lab London Dataset are highlighted by explainable AI approaches such as a Saliency map, CAM, Grad-Cam, and an ensemble of them}
	\label{Face}
	\vspace{4mm}
	\includegraphics[width=0.89\textwidth]{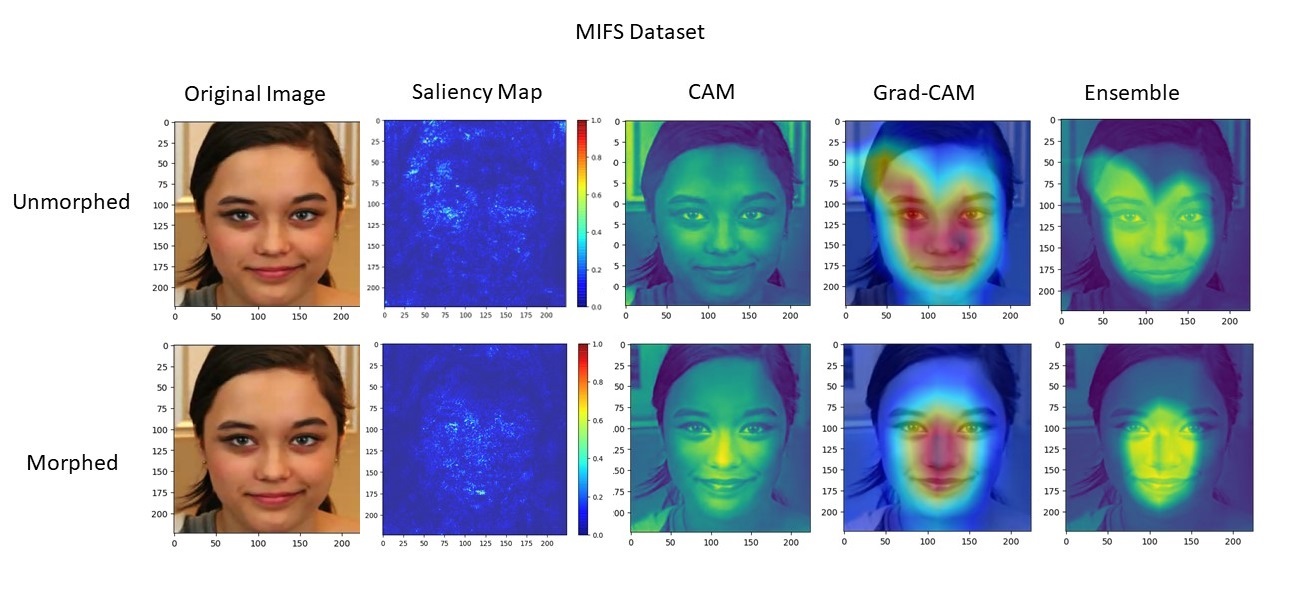}
	\vspace{-6mm}
	\caption{The salient regions for categorising the image as morphed/unmorphed for MIFS Dataset are highlighted by explainable AI approaches such as a Saliency map, CAM, Grad-Cam, and an ensemble of them. }
	\label{MIFS}
	\vspace{4mm}
	\includegraphics[width=0.89\textwidth]{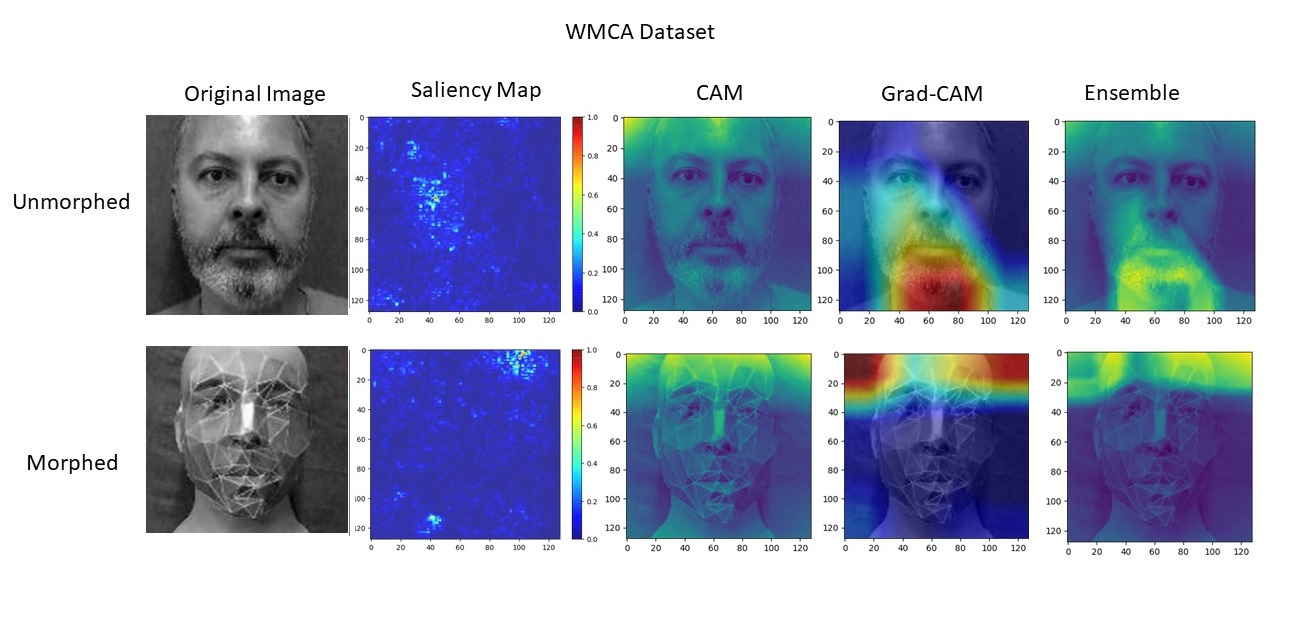}
	\vspace{-6mm}
	\caption{The salient regions for categorising the image as morphed/unmorphed for WMCA Dataset are highlighted by explainable AI approaches such as a Saliency map, CAM, Grad-Cam, and an ensemble of them.}
	\label{WMCA}
\end{figure*}
\end{enumerate}

\subsection{TensorBoard visualization} \label{tensor}

\begin{figure}[htbp]
\includegraphics[width=80mm,scale=0.5]{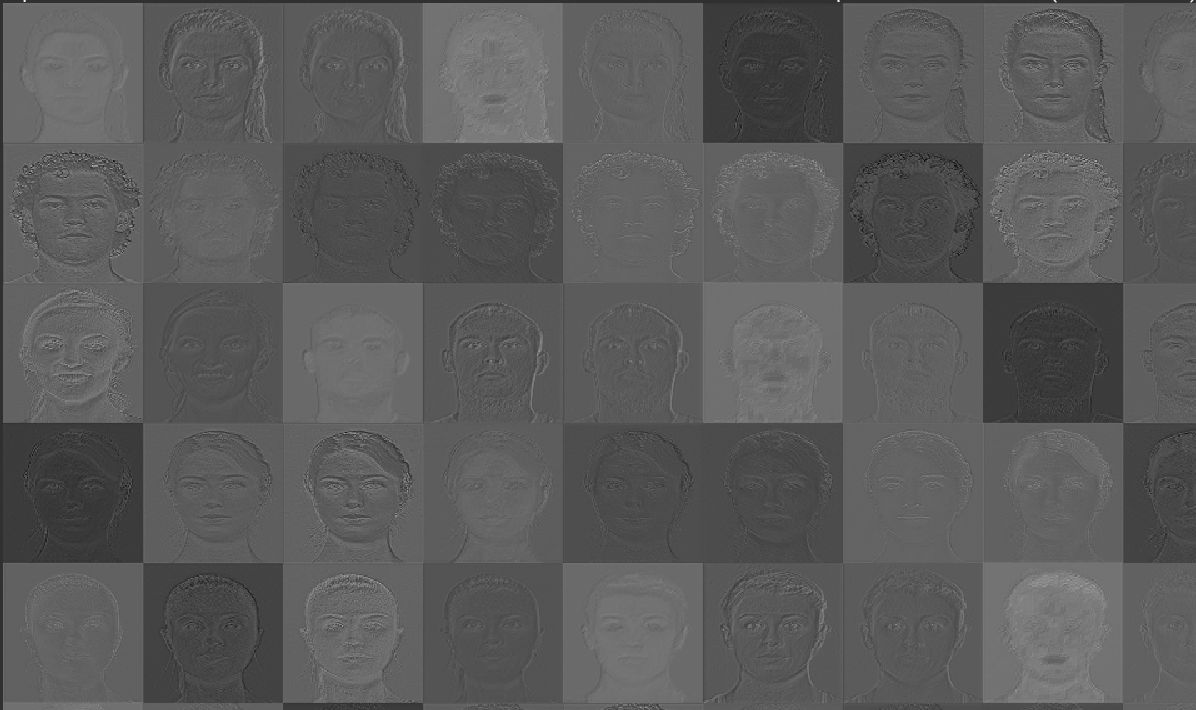}
	\caption{TensorBoard visualization of each layer's output, to determine which layers and weights have contributed most significantly to the model's overall conclusions}
	\label{fig}
\end{figure}

TensorBoard provides the visualizations that are required for neural architectures as it has a rich set of visualizations to track and analyze various indicators such as loss and accuracy. Using TensorBoard, one can easily visualize the structure of a ML model, including the operations and layers used in the model. Additionally, it allows one to track the changes in weights, biases, or other tensors over time using histograms. It also enables users to project embeddings into low-dimensional space for better visualization and analysis. By analyzing the output of each layer, it can determine which layers and weights (between 2 connecting layers) have contributed most significantly to the model's overall conclusions. In Fig. \ref{fig}, the visualization represents the relevant layers and weights over an image.

\section {Conclusion and future scope}
The current work represents a progression towards developing interpretable face biometric PAD solutions to better understand the internal logic of a target model to reach a particular conclusion or decision. We presented a novel stacking-based ensemble approach to find the trade-off between model performance and interpretability. Our results demonstrate that our model is achieving optimal performance on all three datasets and ensemble XAI is generating fine-grained and high-resolution visualizations in support of the model's predictions. As a future scope, emphasis should be laid to develop more innovative biometrics techniques that take interpretability/explainability into account to build trustworthy biometric authentication to be deployed in critical applications such as border access control, airports and military cantonment. Many other interpretability tools can be explored individually or in combination with biometric PAD. Moreover, the evaluation of explanation frameworks is another need of the hour to be addressed. Specifically, in case of face biometric PAD, what level of explanation can be considered as `good' or `accurate'? As a final remark, it is important for the biometrics community to not only evaluate a model using traditional accuracy measures but also to move towards a broader goal of formulating novel performance metrics for explanations.

\bibliographystyle{model2-names.bst}
\bibliography{ref}

\end{document}